\documentclass{article}

\usepackage{microtype}
\usepackage{graphicx}
\usepackage{subfigure}
\usepackage{booktabs} 

\usepackage{nohyperref}[icml2024]

\usepackage[accepted]{icml2024}

\usepackage{amsmath}
\usepackage{amssymb}
\usepackage{mathtools}
\usepackage{amsthm}

\usepackage[capitalize,noabbrev]{cleveref}

\theoremstyle{plain}
\newtheorem{theorem}{Theorem}[section]

\newtheorem{lemma}[theorem]{Lemma}

\theoremstyle{definition}

\theoremstyle{remark}

\usepackage[textsize=tiny]{todonotes}

\icmltitlerunning{Language Alignment via Nash-learning and Adaptive feedback}

\newcommand{\beq}{\begin{equation}}
\newcommand{\eeq}{\end{equation}}

\newcommand{\beqa}{\begin{eqnarray}}
\newcommand{\eeqa}{\end{eqnarray}}

\newcommand{\beqan}{\begin{eqnarray*}}
\newcommand{\eeqan}{\end{eqnarray*}}

\newcommand{\actionspace}{\mathcal{Y}}

\newcommand{\eqdef}{\stackrel{\rm def}{=}}

\usepackage{amsmath,amsfonts,bm}

\def\eqref#1{equation~\ref{#1}}

\def\1{\bm{1}}

\DeclareMathAlphabet{\mathsfit}{\encodingdefault}{\sfdefault}{m}{sl}
\SetMathAlphabet{\mathsfit}{bold}{\encodingdefault}{\sfdefault}{bx}{n}

\newcommand{\KL}{D_{\mathrm{KL}}}

\begin{document}

\twocolumn[
\icmltitle{Language Alignment via Nash-learning and Adaptive feedback}




\begin{icmlauthorlist}
\icmlauthor{Ari Azarafrooz}{comp}
\icmlauthor{Farshid Faal}{comp}
\end{icmlauthorlist}

\icmlaffiliation{comp}{CA, USA}

\icmlcorrespondingauthor{Ari Azarafrooz}{ari.azarafrooz@gmail.com}

\icmlkeywords{Machine Learning, ICML}

\vskip 0.3in
]



\printAffiliationsAndNotice{}  

\begin{abstract}
Recent research has shown the potential of Nash Learning via Human Feedback for large language model alignment by incorporating the notion of a preference model in a minimax game setup.

We take this idea further by casting the alignment as a mirror descent algorithm against the \textit{adaptive feedback} of an \textit{improved opponent}, thereby removing the need for learning a preference model or the existence of an annotated dataset altogether.

The resulting algorithm, which we refer to as \textbf{L}anguage \textbf{A}lignment via \textbf{N}ash-learning and \textbf{A}daptive feedback (LANA), is capable of self-alignment without the need for a human-annotated preference dataset. We support this statement with various experiments and mathematical discussion.

\end{abstract}

\section{Introduction}
The standard approach for Large Language Model (LLM) alignment involves optimizing a reward function that is learned explicitly (RLHF) \cite{christiano2017deep} or implicitly (DPO) \cite{rafailov2024direct} by accessing human-generated feedback. Other alternatives integrate human-generated feedback by learning a preference model that takes two responses, denoted as $y$ and $y'$ (conditioned on a prompt $x$), as input and produces a preference score (a number between 0 and 1), indicating the preference of response $y$ over response $y'$ given the context/prompt $x$. These preference models are then cast as the utility of a game-theoretic framework, leading to the notion of Nash Learning via Human Feedback (NLHF) \cite{munos2023nash}. The solution offered by the Nash equilibrium of the preference model is argued to be more aligned with the diversity of human preferences.

Within a similar minimax game setup, we propose another alternative that uses the \textit{adaptive} feedback of an \textit{improved} opponent without the need for a fixed/learned preference model or pre-generated preference data. This is similar to the transition from RLHF to self-reward DPO \cite{yuan2024self}. \cite{yuan2024self} proposed an offline self-feedback procedure for generating new data for DPO to incorporate into further alignment training. However, no training methodology has yet been proposed to directly incorporate self-evaluating reward functions in the alignment training. This is because self-evaluation could still be noisy and lead to biased and sub-optimal demonstrations. Therefore, learning from such data directly does not guarantee a better optimal model. The main contribution of this paper is to show that self-reward training processes exist that are robust to sub-optimal and noisy iterative self-reward mechanisms. 





\textbf{Comparison with related works}
Two important distinctions of our work compared to related works \cite{munos2023nash,rosset2024direct,yuan2024self,wu2024self} are:

\begin{itemize}
\item All the previous works assume that a preference model is learned in advance, analogous to the concept of reward models in RLHF. However, in our setup we assume we lack access to such a learned preference model or human-annotated preference dataset. Instead, the LLM policies improve through \textbf{\textit{adaptive feedback from improved opponents.}}
\item All the existing works avoid a faithful game-theoretic implementation, such as the two-timescale update, to avoid complex hyperparameter tuning and unstable performance. While this might be true in a generic game-theoretic setup, it seems to be overthinking in the context of LLM alignment. This is because complex policy behaviors are inherently avoidable as a result of a shared common worldview learned in the initial foundation model.
\end{itemize}

Aside from the self-evaluating assumption, our work is also different from \cite{munos2023nash} in that it sets up a modified Mirror Descent algorithm to incorporate the KL regularization with respect to the reference policy. However, our proposed method is reference policy-free.

\cite{chen2024self} proposed self-play in a supervised fine-tuning (SFT) context and not in alignment training.
 
\section{Language Alignment via Nash-learning and Adaptive feedback}

We derive the new alignment algorithm using mirror ascent algorithm with improved opponent (MAIO) \cite{munos2020fast}.
It defines a sequence of policies $(\pi_{i,t})_{t\geq0}$ for a zero-sum game according to the following updates for all $i\in {1,2}$ and for all $t \geq0$:

\begin{eqnarray}\label{eq:maio}
\pi_{i,t+1} = \arg\max_{\pi_i\in\Delta(\actionspace)} [ \gamma_t \pi_{i}.Q_i^{\tilde{\pi}_{-i,t}} - D_\phi(\pi_i,\pi_{i,t})]
\end{eqnarray}

where $\gamma_t$ is a learning rate, $D_{\phi}$ is a Bregman divergence, more specifically a KL distance in our case and $Q_i^{\tilde{\pi}_{-i,t}}$ is the reward of the player $i$ against the improved opponent $\tilde{\pi}_{-i,t}$

\subsection{Sampling from improved Opponent $\tilde{\pi}$}
Improved policies can take different forms, such as greedy, best response, MCTS, extra-gradient method, etc.

The improved policy might also be an optimally aligned model $\pi_{Expert}$. However, we assume that we haven't learned such a model yet. In other words, we haven't trained a preference/reward model in advance, and we are learning this as the game progresses. We hypothesize that one might rely on the self-evaluation of the LLM policy to derive samples from such an improved policy. For example, for a given prompt $x$, we sample two responses $y$ and $y'$.

Each player generates two answers and queries the \textit{opponent} using the following evaluation prompt template:\\
``User\\
Given a piece of instruction and two of its possible responses, output 1 or 2 to indicate which response is better.\\
Instruction: {instruction}, \\
Response 1: {y}\\
Response 2: {y'}?\\
Assistant \\
Preferred response is -''.\\

$y$ is the \textit{preferred} answer for user $i$ if:\\
$\pi_{-i}$(eval~prompt)$[-1][tokenizer("1")]>\\\pi_{-i}$(eval~prompt)$[-1][tokenizer("2")]$ and $y'$ otherwise.

Every player then treats the preferred answer as the sample of an \textit{improved} opponent and the \textit{rejected} answer as their own, aiming to maximize their expected win rate under the setup described in equation \ref{eq:maio} setup.

We show that while this setup leads to noisy outcomes (e.g., the opponent may be wrong) and changing utilities (the preference measure for the exact same response is not identical over the course of the game since the parameters get updated at every step), with the correct choice of proxy reward and slower learning dynamics, the game converges to a better policy.
 

\section{Algorithm}
Let the reward be \textit{adaptively} evaluated at each time $t$ using the policies of the players as follows:
\begin{eqnarray}\label{eq:reward}
Q_i^{\tilde{\pi}_{-i,t}}=\log(\pi_{i,t}/\tilde{\pi}_{-i,t})/\gamma_t
\end{eqnarray}
In section \ref{loss-derivation}, we show that if we plug this into the optimization Eq. \ref{eq:maio}, we end up with \cref{alg:LANA} which we refer to as LANA, short for \textbf{L}anguage \textbf{A}lignment via \textbf{N}ash-learning and \textbf{A}daptive feedback. Note how it can alternatively be viewed as an online, two-player, simplified (reference-free, sigma-free) version of Direct Preference Optimization (DPO) without the need for a human-annotated preference dataset.

\begin{algorithm}[tb]
   \caption{LANA}
   \label{alg:LANA}
\begin{algorithmic}
   \STATE {\bfseries Input:} prompt distribution $\mathcal{X}$, An instruct-tuned LLM model policy $\pi$, eval prompt
    \STATE Initialize $\pi_1,\pi_2\leftarrow \pi$
   \REPEAT
   \STATE $x\sim \mathcal{X}$
   \FOR{$i=1$ {\bfseries to} $2$}
   \STATE $(y,y') \sim \pi_i(x)$
   \STATE Optimize $\pi_i$ using SGD with loss: 
   \IF{$\pi_{-i}(\text{eval prompt})$ indicates $(y' \succ y) $}
   \STATE loss $\coloneqq \log(\pi_{-i}(x,y')/\pi_{i}(x,y))$
   \ELSE
   \STATE loss $\coloneqq \log(\pi_{-i}(x,y)/\pi_{i}(x,y'))$
   \ENDIF
   \ENDFOR
   \UNTIL{Convergence or out of available compute resource}
\end{algorithmic}
\end{algorithm}


\section{Experiments}

\subsection{Experiment Setup}

\subsubsection{Data}
We randomly selected 3K prompts from \cite{DIBT}. We \textbf{only} use the prompts and not the generated responses. For testing aside from common methodology we also utilized pre-processed version of the UltraFeedback test dataset \cite{cui2023ultrafeedback} and different categories of \cite{open_hermes_preferences} for deeper understanding.
 
\subsubsection{Base model}
We used the Phi-3-mini-4k-instruct \cite{abdin2024phi} for most of our experiments. This model is a mini model containing only 3.8b parameters which is helpful for faster experimentation and lack of resources. 

\subsubsection{hyperparameters}
The $y,y'$ were generated using a temperature of $0.1$ with maximum length of 128 tokens. We also limited the prompt to a maximum of 256 tokens. These choices were due to resource limitations. The learning rate for SGD optimization was set to 0.0003, and each batch size was 4. Both players parameterized their base models using LoRA \cite{hu2021lora} with a rank of 16.
 
\subsection{Results}
Alpaca evaluation \cite{alpaca_eval, dubois2024length,dubois2023alpacafarm}
results are shown in Table \ref{alpaca-eval}. It shows noticeable improvement after training using LANA on 3K prompts without access to any human-annotated preference dataset.  

\begin{figure*}[ht]
\vskip 0.2in
\begin{center}
\centerline{\includegraphics[width=2\columnwidth]{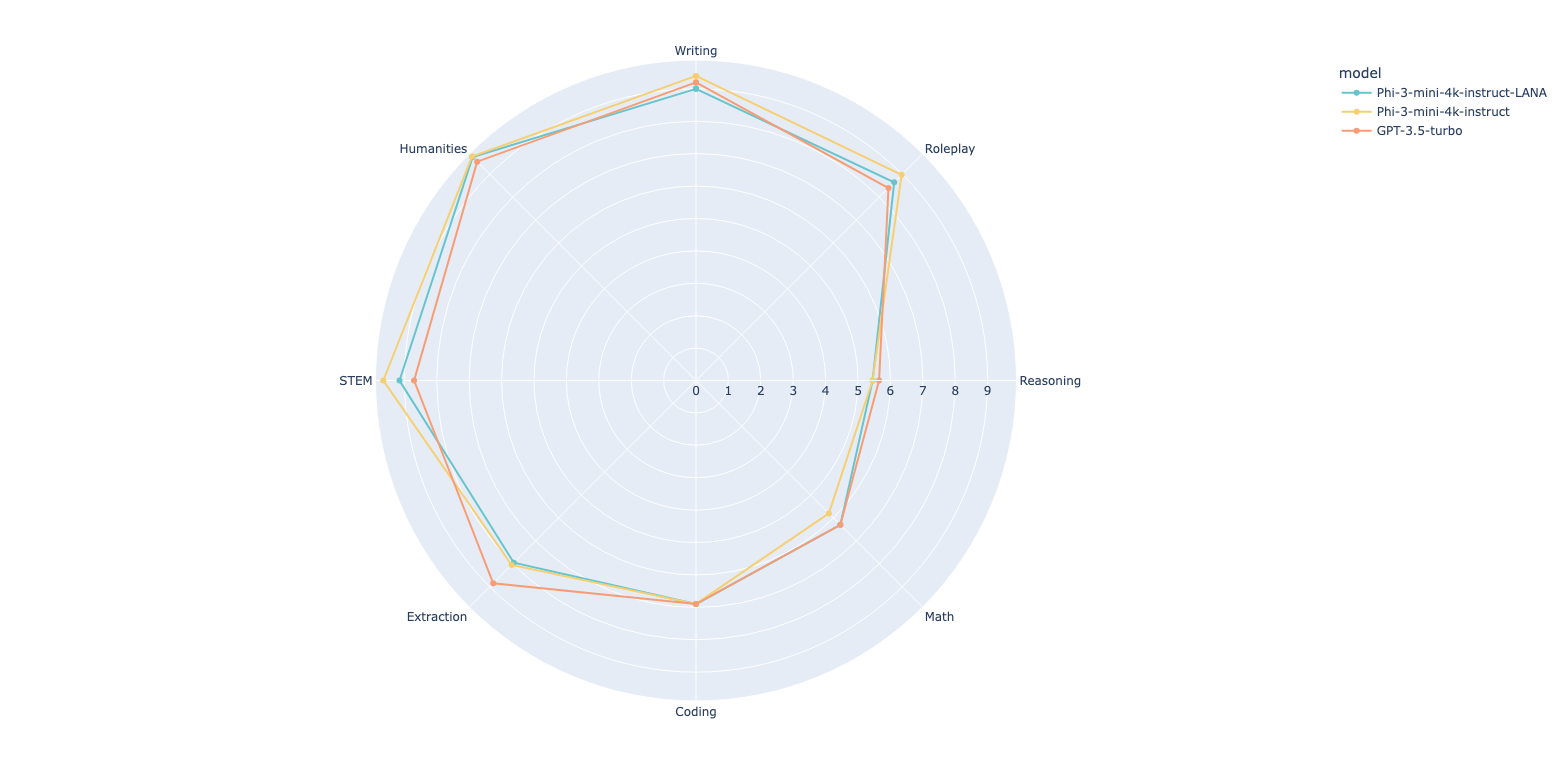}}
\caption{MT-Bench: Unlike other self-rewarding LMs, not only is there no drop in reasoning tasks, but there is also a significant increase in GSM8k, matching that of GPT-3.5-turbo. In other tasks, performance seems to be affected slightly.}
\label{fig:spider}
\end{center}
\vskip -0.2in
\end{figure*}

We also measured the model's performance using LLM-Evaluation Harness \cite{eval-harness}, with results shown in Table \ref{llm-harness-eval} 
and MT-Bench \cite{zheng2023judging}, shown in the spider plot in Fig. \ref{fig:spider}.  The most significant observation is that, unlike other self-reward mechanism such as \cite{yuan2024self}, not only is there no noticeable drop in reasoning tasks, but it also seems to perform noticeably better in  GSM8K task in MT-bench, matching that of GPT-3.5-turbo, despite being a mini model of 3.8b parameters.

\begin{figure*}[ht]
\vskip 0.2in
\begin{center}
\centerline{\includegraphics[width=2\columnwidth]{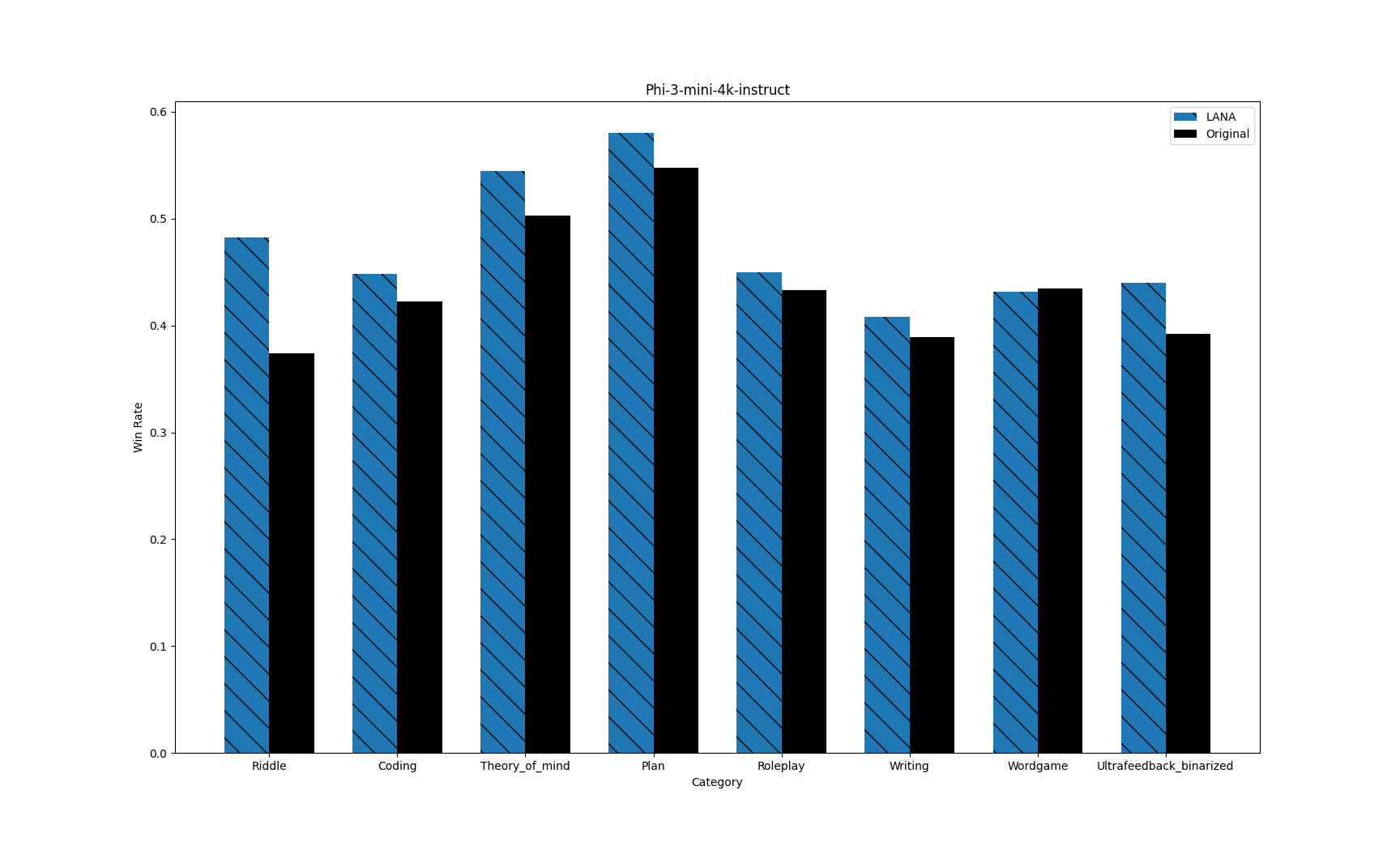}}
\caption{Preference Data set are pre-processed version of the UltraFeedback test dataset \cite{cui2023ultrafeedback} and different categories in \cite{open_hermes_preferences}.}
\label{winrate-phi3}
\end{center}
\vskip -0.2in
\end{figure*}

Next we utilized the annotated preference data set in the  pre-processed version of the UltraFeedback test dataset \cite{cui2023ultrafeedback} and different categories in \cite{open_hermes_preferences} to gain a better understanding across different task categories. The results are demonstrated in Fig. \ref{winrate-phi3}. They show that LANA helped improve the win rate across all tasks, with some  more noticeable improvements in categories such as riddle, theory of mind and plan category, again confirming our assertion regarding the improvements in logical reasoning tasks using LANA without an annotated dataset. 
 
\begin{table}[t]
\caption{Alpca-eval after training on only 3K instructions}
\label{alpaca-eval}
\vskip 0.15in
\begin{center}
\begin{small}
\begin{sc}
\begin{tabular}{lcccr}
\toprule
Model & LC Win Rate & Win Rate & Std  \\
\midrule
LANA   & \textbf{22.50} & \textbf{21.35} & 1.26 \\
Phi-mini   & 20.84 & 19.98 & 1.20  \\

\bottomrule
\end{tabular}
\end{sc}
\end{small}
\end{center}
\vskip -0.1in
\end{table}

\begin{table}[t]
\caption{Unlike other Self-rewarding LM, LANA don't face a significant drop in reasoning tasks.}
\label{llm-harness-eval}
\vskip 0.15in
\begin{center}
\begin{small}
\begin{sc}
\begin{tabular}{lcccr}
\toprule
Task & LANA & BASE \\
\midrule
gsm8k-5shot    & 0.7703& 0.7756 \\
hellaswag & 0.7822& 0.7841 \\
arc-challenge    & 0.5674 & 0.5785 \\
hellaswag    & 0.7822 & 0.7841 \\

\bottomrule
\end{tabular}
\end{sc}
\end{small}
\end{center}
\vskip -0.1in
\end{table}

\subsection{Ablation study}

\begin{figure*}[ht]
\vskip 0.2in
\begin{center}
\centerline{\includegraphics[width=2\columnwidth]{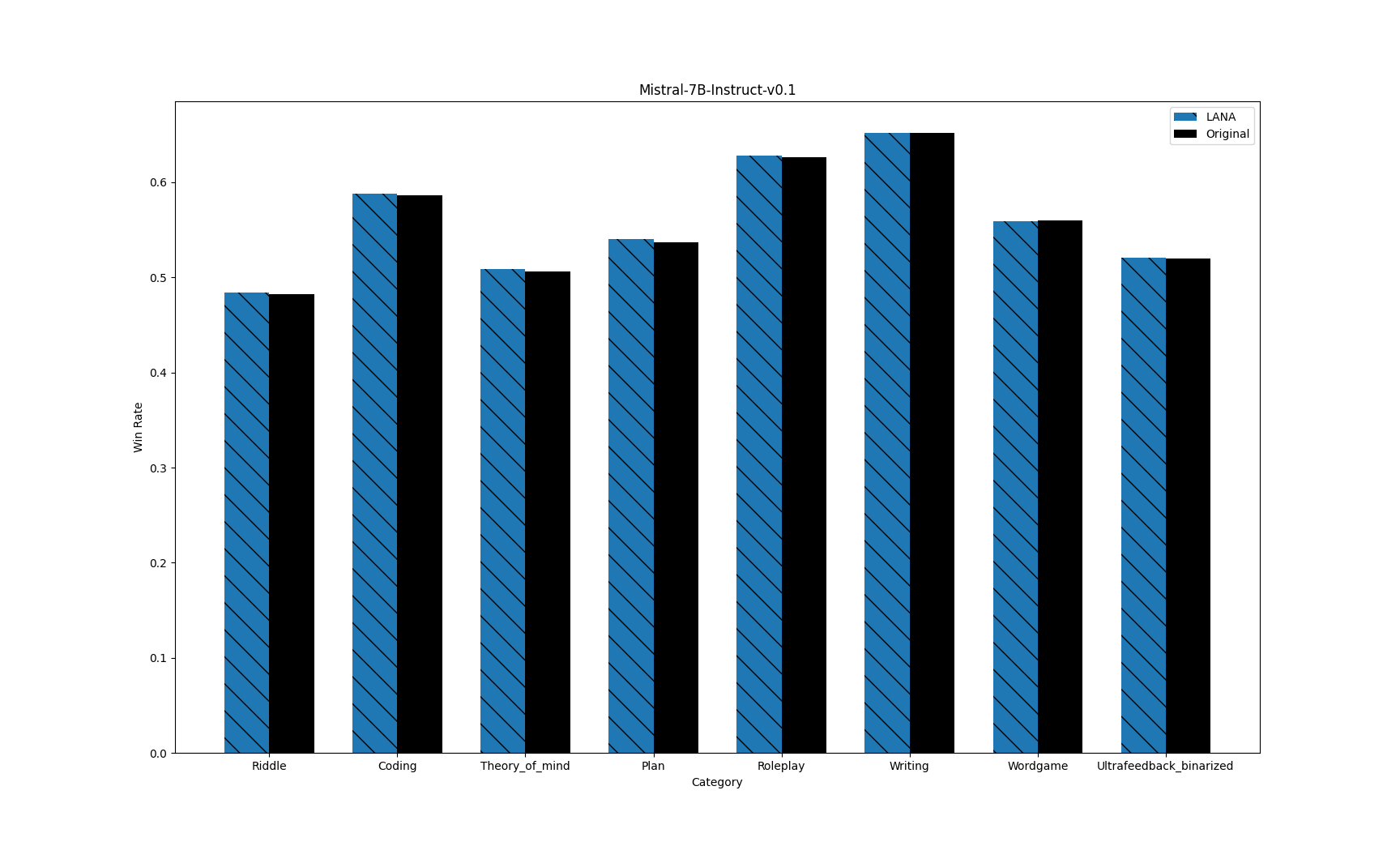}}
\caption{\textbf{LANA provides no benefit with Mistral-v1 as foundation model}. Preference datasets are pre-processed version of the UltraFeedback test dataset \cite{cui2023ultrafeedback} and different categories in \cite{open_hermes_preferences}.}
\label{winrate-mistral}
\end{center}
\vskip -0.2in
\end{figure*}

We tested two additional models, Mistral-7B-Instruct-v0.1 \cite{jiang2023mistral} and Gemma-2b-it \cite{team2024gemma}. Both led to negligible improvements as shown in Fig. \ref{winrate-mistral}. While we cannot rule out other reasons, we suspect this implies that the choice of the base model is crucially important. This could mean that the lower quality of training data in these models leads to noisier self-evaluations, which in turn seem to cancel out the progress made during alignment. In contrast, the Phi model training data appears to be collected using the "TextBook all you need" methodology \cite{gunasekar2023textbooks}. 

\section{Mathematical Discussions}

Two points need to be addressed:

\begin{itemize}
\item How LANA loss function is derived from Eq. \ref{eq:maio}?
\item Does the game converge, and it what sense?
\end{itemize}

\subsection{LANA loss derivation}\label{loss-derivation}
An alternative way to define Eq. \ref{eq:maio} is through mirror maps  \cite{bubeck2015convex}. A mirror map is a mapping induced by the convex function $\phi$, which maps primal variables to dual variables. Given a convex function $\phi$, the mirror map is essentially the gradient $\nabla_{\phi}$. For the KL divergence case, the mirror map associated with the negative entropy function is given by the following gradients:
\begin{eqnarray}\label{eq:mirror-map} 
\nabla_{\phi(\pi)}=\log(\pi)
\end{eqnarray}

Then the alternative definition for Eq. \ref{eq:maio} is:
\begin{eqnarray}\label{eq:dual}
\pi_{i,t+1} = \arg\min_{\pi_i\in\Delta(\actionspace)} D_\phi(\pi_i,z_{i,t+1})
\end{eqnarray}

where $z_{i,t+1}$ is such that:

\begin{eqnarray}\label{eq:zdef}
\nabla_{\phi(z_{i,t+1})}=\nabla_{\phi(\pi_{i,t})} + \gamma Q_i^{\tilde{\pi}_{-i,t}}
\end{eqnarray}

In other words, gradient descent steps are performed in mirror space (policy log-likelihood) instead of in LLM weights space.

Combining the above equations with reward Eq.\ref{eq:reward} yields:

 \begin{eqnarray}\label{eq:duals-simplified}
\pi_{i,t+1} = \arg\min_{\pi_i\in\Delta(\actionspace)}\mathop{\mathbb{E}_{\pi_i}}[\log (\tilde{\pi}_{-i,t}/\pi_{i,t})] - H(\pi_i)
\end{eqnarray}

We instead minimize the upper bound by ignoring the entropy term:

\begin{eqnarray}\label{eq:lana-loss}
loss \coloneqq  \mathop{\mathbb{E}_{\pi_i}} \log (\tilde{\pi}_{-i,t}/\pi_{i,t})
\end{eqnarray}
which we pass as the loss to SGD for optimization.

We also note that the choice of $Q$ is essential in deriving such a bound, and it is not simply the result of ignoring the KL term in \eqref{eq:maio}. The KL term in Eq. \eqref{eq:maio} $-\KL(\pi,\pi_{i,t})=H(\pi)-H(\pi,\pi_{i,t})$ has two goals: first, to encourage exploration for $\pi$ via the entropy maximization term $H(\pi)$; and second, to avoid reward hacking so that $\pi$ doesn't deviate too much from the past policy $\pi_{i,t}$ by minimizing the cross-entropy $H(\pi,\pi_{i,t})$. The cross-entropy term is still captured in the objective of Eq. \ref{eq:lana-loss}. However, our experiments show that entropy term is not important, and therefore for the sake of simplicity is removed from the loss term.
\subsection{Convergence }
Let's drop the player index notation as the game is symmetric.





\begin{lemma} \cite{munos2020fast}
Let $p\geq 1$ and $q\geq 1$ such that $1/p+1/q=1$. Let $\phi$ be a strongly convex function with respect to the $\ell_p$-norm $\|\cdot \|_p$ with some modulus $\sigma$, i.e., for any $\pi,\pi'$,
\beq \notag
\phi(\pi)\geq \phi(\pi') + \nabla \phi(\pi')\cdot(\pi-\pi')+\frac{\sigma}{2} \|\pi-\pi'\|^2.
\eeq
Write $D_\phi$ the associated Bregman divergence: for $\pi,\pi'$,
$$D_\phi(\pi,\pi')\eqdef \phi(\pi)-\phi(\pi') - \nabla \phi(\pi')\cdot( \pi-\pi').$$
Let $\delta$ be a vector of dimension $|\actionspace|$. Define $\pi_{t+1}$ as
\beq
\pi_{t+1} = \arg\max_{\pi\in\Delta(\actionspace)} \left\lbrack   \pi.\delta_t - D_\phi(\pi,\pi_t) \right\rbrack,
\eeq
Then for any $\pi\in \Delta(\actionspace)$, we have, 
\beqan 
D_\phi(\pi,\pi_{t+1}) \leq D_\phi(\pi, \pi_{t}) +  (\pi_t-\pi).\delta_t + (2/\sigma) \|\delta_t\|_q^2.
\eeqan
\end{lemma}

Using the last lemma with the choice  of $Q$ in eq. \ref{eq:reward} and $D_\phi$ being a $\KL$ distance, we have

$\delta_t=\gamma_t\log(\tilde{\pi}_t/\pi_{t})$ 

It follows that

\begin{eqnarray}\label{eq:convergence-ineqality}
\KL(\pi,\pi_{t+1}) \leq  \notag \\ \KL(\pi, \pi_{t}) +  (\pi-\pi_t).\gamma_t\log(\pi_{t}/\tilde{\pi_t})\ + (2/\sigma) \|\delta_t\|_q^2 \notag \\ 
\leq  \KL(\pi, \pi_{t}) +  (\pi.\gamma_t\log(\pi_{t}/\tilde{\pi_t})\ + (2/\sigma) \|\delta_t\|_q^2 \notag \\ 
\leq  (1-\gamma_t) \KL(\pi, \pi_{t})  + \gamma_t \KL(\pi,\tilde{\pi_t}) + (2/\sigma) \|\delta_t\|_q^2 
\end{eqnarray}
where the second inequality is the result of $KL(\pi_t,\tilde{\pi_t})>0$
and the last inequality is the result of re-writing $\log(\pi_{t}/\tilde{\pi_t})=\log(\pi/\tilde{\pi_t})*\log(\pi_{t}/\pi)$ 

By iterating the inequality and assuming the norm is bounded (for example, by ensuring that the policy probability does not have zero support), we can make the following conclusion for a fixed learning rate $\gamma_t=\gamma$:

\begin{eqnarray}\label{eq:convergence}
\KL(\pi^*,\pi_{T}) \leq  \notag \\ \KL(\pi^*,\tilde{\pi_c}) +  (1-\gamma)^T \KL(\pi^*, \pi_{0}) + (2/\gamma\sigma) \|\delta_t\|_q^2 \notag \\
\leq  \KL(\pi^*,\tilde{\pi_c}) + e^{-\gamma T} \KL(\pi^*, \pi_{0}) + (2/\gamma\sigma) \|\delta_t\|_q^2 
\end{eqnarray}

where $\pi^*$ is Nash equilibrium, $c=\arg\max_{t\in {0,...,T}}\KL(\pi^*,\tilde{\pi_t})$. From it, we can conclude convergence on average but last iterate convergence is not guaranteed (unless $\KL(\pi^*,\tilde{\pi_c})=0$).










\subsection{Future works}
An important future direction is to test LANA across different and larger models to assess the role of the base model. Additionally, we didn't have enough compute to train on more than 3k prompts. The question remains how much more improvement LANA alignment would have provided with continued training on more instructions.

Moreover, sampling during training leads to highly inefficient training. Tricks such as PagedAttention \cite{kwon2023efficient} are also not applicable during training. Improving the training efficiency of LANA is another important area for future work.







\bibliography{example_paper}
\bibliographystyle{icml2024}




\end{document}